# Fuzzy and Multilayer Perceptron for Evaluation of HV Bushings

Sizwe M. Dhlamini, Tshilidzi Marwala, and Thokozani Majozi

*Abstract*—**The work proposes the application of fuzzy set theory (FST) to diagnose the condition of high voltage bushings. The diagnosis uses dissolved gas analysis (DGA) data from bushings based on IEC60599 and IEEE C57-104 criteria for oil impregnated paper (OIP) bushings. FST and neural networks are compared in terms of accuracy and computational efficiency. Both FST and NN simulations were able to diagnose the bushings condition with 10% error. By using fuzzy theory, the maintenance department can classify bushings and know the extent of degradation in the component.**

## I. Introduction

THIS work presents fuzzy set theory (FST) used in condition monitoring for high voltage bushings. Fuzzy set theory (FST) has been used in diverse applications in the last decade, Majozi and Zhu [1] used FIS to match operators and chemical plants based on their skill, availability, health and age. Kubica, Wang and Winter [2], used FST in control systems; Flaig, Barner and Arce [3] applied FST in pattern recognition. Ammar and Wright [4] applied FST in the evaluation of state government performance, client satisfaction surveys, and economic impact of state-funded agencies. Its main strength is the ability to model imprecise or uncertain data that characterises many systems and environments. Fuzzy theory allows one to explore the interaction of variables which define a system, and how the variables affect the system's output. Majozi [1] emphasises that attempting to linearly combine these inputs would not be able to explore these interactions, hence would lack robustness. Neural Networks have been tested by Dhlamini and Marwala for condition monitoring of bushings [5] and by Wang [6] for transformer condition monitoring. In the case of bushings that are evaluated using IEC60599 [12] there is a large range of values associated with normal, elevated and abnormal amounts of gas. Fuzzy will help in objectively answering the question: How high is too high or too low for an elevated condition to be classified as dangerous and require the bushing to be maintained.

S. M. Dhlamini is with Eskom, Distribution Technology, Private Bag x1074, Germiston, 1400, South Africa, sizwe.dhlamini@eskom.co.za

T. Marwala is Professor at the School of Electrical and Information Engineering University of the Witwatersrand, Private Bag x3, Johannesburg, 2000, South Africa, t.marwala@ee.wits.ac.za.

T. Majozi is Associate Professor at the Department of Chemical Engineering, University of Pretoria, 0002, South Africa, thoko.majozi@up.ac.za

### *A. Background*

There are 4 steps involved in fuzzy logic implementation, i.e. 1) Fuzzify inputs, 2) Select membership functions, 3) Apply fuzzy operators, and finally 4) Defuzzify [1].

1.1.1 **Fuzzify** inputs means: to identify the inputs or attributes which describe the system.

1.1.2 Select **membership functions** means: to resolve all fuzzy statements (inputs) into a degree of membership between 0 and 1 for each attribute.

1.1.3 Apply **fuzzy operators** means: to AND or OR or NOT or ANY the inputs similarly to Boolean algebra. AND is the ***min*** fuzzy operator, chooses the least of all values inputs in the same rule. OR is the ***max*** fuzzy operator, chooses the greatest of all values input in the same rule. The ***not*** operator makes the value the opposite, i.e. (1-value). The ***any*** operator sums the values in the rule. The result after applying the fuzzy operator is called the degree of support for the rule. e.g. max (0.0, 0.7) = 0.7 means if inputs are 0.0 or 0.7 then choose 0.7. Fuzzy sets need more than one rule, Majozi and Zhu [1] generally used three rules.

1.1.4 **Defuzzify** means: to apply the implication or consequence. This is done by using the degree of support for the entire rule to shape the fuzzy set output.

Mamdani [7] and Sugeno [8] proposed two types of fuzzy inference systems (FIS) that are commonly used. The more popular of the two is Mamdani fuzzy inference (MFI), first proposed by Ebrahim Mamdani in 1975. He used the method to control a steam engine boiler by using linguistic control rules from experienced human operators in a machine controlled system. Mamdani based that work on Lofti Zadeh's work [9] which was published in 1973. Mamdani fuzzy inference expects the membership function to be fuzzy sets. After summation the output must be defuzzified. MFI finds the centroid of a 2D function. It uses a single output membership function because it greatly simplifies computation of MFI. Rather than integrate across the entire 2D function to find the centroid, MFI uses weighted average of a few data points. Sugeno fuzzy inference is normally used to model systems where the output is linear or constant.

## II. FST for Bushing Evaluation

Fuzzy set theory is used to explore the interrelation between each bushing's identifying attributes, i.e. the dissolved gases in oil. In dissolved gas analysis (DGA) there is a relation between consequent failure and the simultaneous presence of oxygen with a secondary gas such as hydrogen, methane, ethane, ethylene, acetylene, and carbon monoxide in a bushing. The presence of combustible gasses in the absence of oxygen is itself not an indication of eminent failure. Applying fuzzy sets on bushing data is necessary because the extent to which the evaluation criterion is below the threshold for a safe and acceptable or rejected due risk of explosion, is not uniform for each bushing. This discrepancy can be accounted for in the evaluation process by applying fuzzy set theory. Temperature is an important criterion in the evaluation. Temperature refers both to the operating temperature of the oil and the difference between ambient and the oil temperature. Bushings that continuously operate at temperatures near or above the auto-ignition temperature of any of the gases or oil have a significantly higher probability of explosion than those that operate at lower temperatures with the same ratio of gases. The American Society for Testing and Materials (ASTM), document ASTM D 2155 defines the auto-ignition temperature of a substance is the temperature at or above which a material will spontaneously ignite or catch fire without an external spark or flame [10].

Auto-ignition temperature should not to be confused with flash or fire points, which are generally a few hundred degrees lower. The flash point is the lowest temperature at which a liquid can form an ignitable mixture with air near the surface of the liquid. The lower the flash point, the easier it is to ignite the material. Fire point is the minimum sample temperature at which vapour is produced at a sufficient rate to sustain combustion. It is the lowest temperature at which the ignited vapour persists in burning for at least 5 seconds.

Flash point may be determined by the ASTM D 93 Method called "Flash Point by Pensky-Martens Closed Tester" for fuel oils. Alternatively ASTM D 92 Method called "Flash and Fire Points by Cleveland Open Cup" can determine flash points of lubricating oils. At the fire point, the temperature of the flame becomes self-sustained so as to continue burning the liquid, while at the flash point; the flame does not need to be sustained.

The fire point is usually a few degrees above the flash point. Transformer oil which is used for both cooling and electrical insulation has characteristics as shown in Table 1.

## A. Identifying Attributes

In this study ten identifying attributes were selected to develop membership functions. These are concentrations of hydrogen, oxygen, nitrogen, methane, carbon monoxide, carbon dioxide, ethylene, ethane, acetylene and total dissolved combustibles gases. The concentrations are in parts per million (ppm). IEC60599 and IEEE C57-104 criteria were used in decision making.

TABLE I
PROPERTIES OF BUSHING OIL

| Property | Magnitude |
|---|---|
| Boiling point | 140ºC (at 10mmHg) |
| Vapor Pressure | 0.1 mbar (10Pa) [at 20ºC] |
| Density | 840 kg/m$^3$ at 15ºC |
| Specific Gravity | 0.8890 |
| Solubility in $H_2O$ | insoluble |
| Viscosity | 7.7 mm at 40ºC |
| Flash point | 156ºC |
| Auto-ignition Temperature | 250ºC |

## B. Membership Functions

Defining the membership functions (MF) is the most important step in fuzzy set theory application. This step takes the most time and must be accurate. One can use other MF curves such as a straight-line, Gaussian-bell, sigmoid, polynomial or a combination, if one can justify the decision after analysis of the data. Bojadziev and Bojadziev [14] discussed that triangular functions accurately represent most memberships. In general, triangular and trapezoidal membership functions are representative of most cases [1], [9]. In this application the trapezoidal and triangular shapes of membership functions were selected to coincide with the safe operating limits for gas contaminants inside the bushings oil. Each of the attributes is rated in terms of high, medium or low. The rating depends on the measured magnitude of the attribute compared to the reject threshold obtained in IEC60599 criteria. The membership functions (MF) are as given in Equations (1) to (31).

**Hydrogen**

$$\mu_{Normal}(x) = \begin{cases} 1 \cdots 0 \le x \le 135 \\ -0.067x + 10 \cdots 135 \le x \le 150 \end{cases} \quad (1)$$

$$\mu_{\text{Elevated}}(x) = \begin{cases} 0.067x - 9 \cdots 135 \le x \le 150 \\ 1 \cdots 150 \le x \le 900 \\ -0.067x + 10 \cdots 900 \le x \le 1000 \end{cases} \quad (2)$$

$$\mu_{\text{Dangerous}}(x) = \begin{cases} 0.01x - 9 \cdots 900 \le x \le 1000 \\ 1 \cdots x \ge 1000 \end{cases} \quad (3)$$

**Methane**

$$\mu_{Normal}(x) = \begin{cases} 1 \cdots 0 \le x \le 23 \\ -0.5x + 12.5 \cdots 23 \le x \le 25 \end{cases} \quad (4)$$

$$\mu_{\text{Elevated}}(x) = \begin{cases} 0.5x - 11.5 \cdots 23 \le x \le 25 \\ 1 \cdots 25 \le x \le 72 \\ -0.125x + 10 \cdots 72 \le x \le 80 \end{cases} \quad (5)$$

$$\mu_{\text{Dangerous}}(x) = \begin{cases} 0.125x - 9 \cdots 72 \le x \le 80 \\ 1 \cdots x \ge 80 \end{cases} \quad (6)$$

**Ethane**

$$\mu_{Normal}(x) = \begin{cases} 1 \cdots 0 \le x \le 9 \\ -x + 10 \cdots 9 \le x \le 10 \end{cases} \quad (7)$$

$$\mu_{\text{Elevated}}(x) = \begin{cases} x - 9 \cdots 9 \le x \le 10 \\ 1 \cdots 10 \le x \le 32 \\ -0.333x + 11.66 \cdots 32 \le x \le 35 \end{cases} \quad (8)$$

$$\mu_{\text{Dangerous}}(x) = \begin{cases} 0.333x - 10.66 \cdots 32 \le x \le 35 \\ 1 \cdots x \ge 35 \end{cases} \quad (9)$$

**Ethylene**

$$\mu_{Normal}(x) = \begin{cases} 1 \cdots 0 \le x \le 18 \\ -0.5x + 10 \cdots 18 \le x \le 20 \end{cases} \quad (10)$$

$$\mu_{\text{Elevated}}(x) = \begin{cases} 0.5x - 9 \cdots 18 \le x \le 20 \\ 1 \cdots 20 \le x \le 90 \\ -0.1x + 10 \cdots 90 \le x \le 100 \end{cases} \quad (11)$$

$$\mu_{\text{Dangerous}}(x) = \begin{cases} 0.1x - 9 \cdots 90 \le x \le 100 \\ 1 \cdots x \ge 100 \end{cases} \quad (12)$$

**Acetylene**

$$\mu_{Normal}(x)=\begin{cases}1\cdots 0\le x\le 14\\ -x+15\cdots 14\le x\le 15\end{cases} \quad (13)$$

$$\mu_{\text{Elevated}}(x)=\begin{cases}x-14\cdots 14\le x\le 15\\ 1\cdots 15\le x\le 63\\ -0.142857x+10\cdots 63\le x\le 70\end{cases} \quad (14)$$

$$\mu_{\text{Dangerous}}(x)=\begin{cases}0.142857x-9\cdots 63\le x\le 70\\ 1\cdots x\ge 70\end{cases} \quad (15)$$

**Carbon Monoxide**

$$\mu_{Normal}(x)=\begin{cases}1\cdots 0\le x\le 450\\ -0.02x+10\cdots 450\le x\le 500\end{cases} \quad (16)$$

$$\mu_{\text{Elevated}}(x)=\begin{cases}0.02x-9\cdots 450\le x\le 500\\ 1\cdots 500\le x\le 900\\ -0.01x+10\cdots 900\le x\le 1000\end{cases} \quad (18)$$

$$\mu_{\text{Dangerous}}(x)=\begin{cases}0.01x-9\cdots 900\le x\le 1000\\ 1\cdots x\ge 1000\end{cases} \quad (19)$$

**Nitrogen**

$$\mu_{Normal}(x)=\begin{cases}1\cdots 0\le x\le 0.9\\ -10x+10\cdots 0.9\le x\le 1\end{cases} \quad (20)$$

$$\mu_{\text{Elevated}}(x)=\begin{cases}10x-9\cdots 0.9\le x\le 1\\ 1\cdots 1\le x\le 9\\ -x+10\cdots 9\le x\le 10\end{cases} \quad (21)$$

$$\mu_{\text{Dangerous}}(x)=\begin{cases}x-9\cdots 9\le x\le 10\\ 1\cdots x\ge 10\end{cases} \quad (22)$$

**Oxygen**

$$\mu_{Normal}(x)=\begin{cases}1\cdots 0\le x\le 0.09\\ -100x+10\cdots 0.09\le x\le 0.1\end{cases} \quad (23)$$

$$\mu_{\text{Elevated}}(x)=\begin{cases}100x-9\cdots 0.09\le x\le 0.10\\ 1\cdots 0.10\le x\le 0.18\\ -50x+10\cdots 0.18\le x\le 0.20\end{cases} \quad (24)$$

$$\mu_{\text{Dangerous}}(x)=\begin{cases}50x-9\cdots 0.18\le x\le 0.20\\ 1\cdots x\ge 0.20\end{cases} \quad (25)$$

**Carbon Dioxide**

$$\mu_{Normal}(x)=\begin{cases}1\cdots 0\le x\le 9000\\ -0.001x+10\cdots 9000\le x\le 10000\end{cases} \quad (26)$$

$$\mu_{\text{Elevated}}(x)=\begin{cases}0.001x-9\cdots 9000\le x\le 10000\\ 1\cdots 10000\le x\le 13500\\ -0.00067x+10\cdots 13500\le x\le 15000\end{cases} \quad (27)$$

$$\mu_{\text{Dangerous}}(x)=\begin{cases}0.00067x-9\cdots 13500\le x\le 15000\\ 1\cdots x\ge 15000\end{cases} \quad (28)$$

**Total Combustible Gases**

$$\mu_{Normal}(x)=\begin{cases}1\cdots 0\le x\le 648\\ -0.01389x+10\cdots 648\le x\le 720\end{cases} \quad (29)$$

$$\mu_{\text{Elevated}}(x)=\begin{cases}0.01389x-9\cdots 648\le x\le 720\\ 1\cdots 720\le x\le 4500\\ -0.002x+10\cdots 4500\le x\le 5000\end{cases} \quad (30)$$

$$\mu_{\text{Dangerous}}(x)=\begin{cases}0.002x-9\cdots 4500\le x\le 5000\\ 1\cdots x\ge 5000\end{cases} \quad (31)$$

### *C. Fuzzy Rules*

Fuzzy rules represent interrelation between all the inputs. The number of rules is theoretically equal to the number of fuzzy categories raised to the power of the number of fuzzy criteria. Fuzzy categories (FC) used in this case were, the membership functions “dangerous”, “elevated” or “normal”. Fuzzy criteria (NC) used in this case were the different gases that are present, i.e. hydrogen, methane, ethane, ethylene, acetylene, carbon monoxide, nitrogen, oxygen, carbon dioxide and total combustible gases. The rates of change of the gases were not used because the available data is taken on one day only. The required number of fuzzy rules is calculated according to Eq 32. Rules have an antecedent and a consequence. Rules can be expressed in the form:

IF Attribute1 is $A_1$ AND Attribute2 is $A_2$ AND …AND AttributeN is $A_N$, THEN Consequent is Ci,

In the expression, Attribute1, Attribute2,.., AttributeN collectively form an Antecedent. Antecedents and Consequents are variables or concepts and A1, A2; …, Ci are linguistic terms or fuzzy sets of these variables, such as, “low”, “dangerous” or “high”, etc..

$$N_{rules}=(FC)^{NC} \quad (32)$$

$$N_{rules}=(3)^{10}=59049 \quad (33)$$

### *D. Simplification of Fuzzy Rules*

In the case of bushing diagnosis the combinations of the combustible gases in the absence of oxygen does not create a failure. With transformer oil, failure occurs when oxygen is present in quantities above 0.2% at temperatures above 250ºC without any spark present (auto-ignition) or at 156ºC if a spark is present (flash point). This condition reduces the number of fuzzy rules significantly, to only of 81 fuzzy rules. The combinations are modelled in 24 compartments shown in Table 2. Two examples of fuzzy rules in spoken language (as opposed to machine language) are 1) If hydrogen is High only then Low Risk and 2) If hydrogen is High and Oxygen is High then High Risk.

### *E. Consequence or Decision Table*

Based on the rules the bushing is given a risk rating for which certain maintenance actions must be taken on the plant. For safe operation of bushings it is recommended that all HR cases, trip the transformer and remove the bushing from the transformer. For all MR cases monitor the bushings more frequently, i.e. reduce the sampling interval by half. All LR cases operate as normal. From the decision table an aggregated membership is developed, shown in Equations 34 and 35

$$\mu_{agg}=\mu_{LR}\cup\mu_{MR}\cup\mu_{HR} \quad (34)$$

Where

$$\mu_{LR}(x)=\begin{cases}1\cdots x\leq 10\\ -0.01667x+1\cdots 10\leq x\leq 60\end{cases}$$

$$\mu_{MR}(x)=\begin{cases}0.01667x-1\cdots 10\leq x\leq 60\\ -0.05x+4\cdots 60\leq x\leq 80\end{cases} \qquad (35)$$

$$\mu_{HR}(x)=\begin{cases}0.05x-3\cdots 60\leq x\leq 80\\ 1\cdots x\geq 80\end{cases}$$

The decision table and a graph of the membership functions are shown in Table 3 and Figure 1 respectively. The conclusion table shows values and classes. The values of 10, 60 and 80 were selected to represent the levels of risk of failure of a bushing. These levels were then taken as the limits of each of the groups in the conclusion membership function.

The membership function is asymmetrical so that a decision to exclude damaged bushings is more stringent than that of marginally safe bushings. In other words, small changes in a condition that is becoming dangerous are highlighted by the membership function. A steeper gradient on the graph allows the user to identify those components which have small differences in dangerous levels of concentrations of dangerous gases

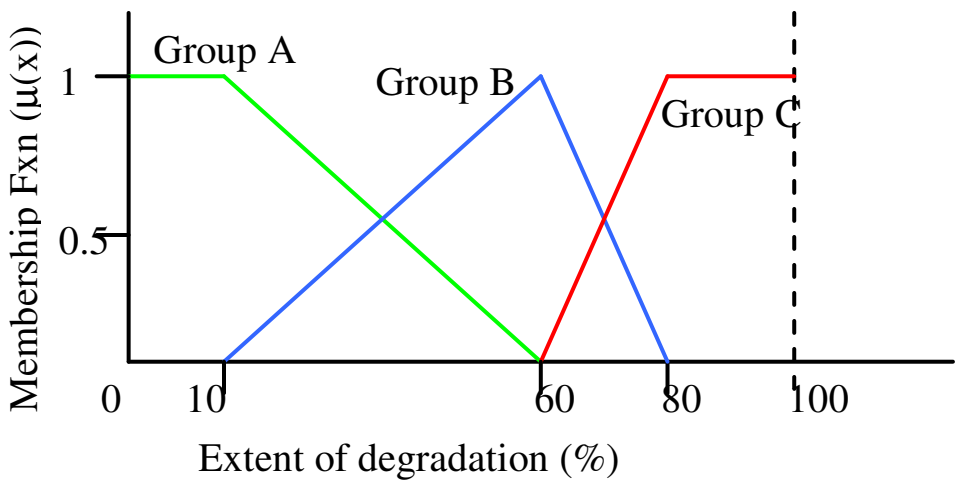

Fig. 1. Membership functions of decision

TABLE III
CONCLUSION TABLE

| Conclusion Table | | | |
|---|---|---|---|
| x | LR | MR | HR |
| 0 | 1 | | |
| 10 | 1 | 0 | |
| 60 | 0 | 1 | 0 |
| 80 | | 0 | 1 |
| 100 | | | 1 |
| | Group A | Group B | Group C |

TABLE II
FUZZY DECISION TABLE

| Oxygen | Total Dissolved Combustible Gases (TDCG) | | | | | | | | | | | |
|---|---|---|---|---|---|---|---|---|---|---|---|---|
| | Dangerous | | | | | | | | | | | |
| | Hydrogen | | | Methane | | | Ethane | | | Acetylene | | |
| | Dangerous | Elevated | Normal | Dangerous | Elevated | Normal | Dangerous | Elevated | Normal | Dangerous | Elevated | Normal |
| Dangerous | HR | HR | MR | HR | HR | MR | HR | HR | MR | HR | HR | MR |
| Elevated | HR | HR | MR | HR | HR | MR | HR | HR | MR | HR | HR | MR |
| Normal | MR | MR | LR | MR | MR | LR | MR | MR | LR | MR | MR | LR |

| Oxygen | Total Dissolved Combustible Gases (TDCG) | | | | | | | | | | | |
|---|---|---|---|---|---|---|---|---|---|---|---|---|
| | Dangerous | | | | | | | | | | | |
| | Carbon Monoxide | | | Nitrogen | | | Carbon Dioxide | | | Ethylene | | |
| | Dangerous | Elevated | Normal | Dangerous | Elevated | Normal | Dangerous | Elevated | Normal | Dangerous | Elevated | Normal |
| Dangerous | HR | HR | MR | MR | HR | HR | MR | MR | MR | HR | HR | MR |
| Elevated | HR | HR | MR | MR | HR | HR | MR | MR | LR | HR | HR | MR |
| Normal | MR | MR | LR | LR | LR | LR | LR | LR | LR | MR | MR | LR |

| Oxygen | Total Dissolved Combustible Gases (TDCG) | | | | | | | | | | | |
|---|---|---|---|---|---|---|---|---|---|---|---|---|
| | Elevated | | | | | | | | | | | |
| | Hydrogen | | | Methane | | | Ethane | | | Acetylene | | |
| | Dangerous | Elevated | Normal | Dangerous | Elevated | Normal | Dangerous | Elevated | Normal | Dangerous | Elevated | Normal |
| Dangerous | HR | HR | MR | HR | HR | MR | HR | HR | MR | HR | HR | MR |
| Elevated | HR | HR | MR | HR | HR | MR | HR | HR | MR | HR | HR | MR |
| Normal | MR | LR | LR | MR | LR | LR | MR | LR | LR | MR | LR | LR |

| Oxygen | Total Dissolved Combustible Gases (TDCG) | | | | | | | | | | | |
|---|---|---|---|---|---|---|---|---|---|---|---|---|
| | Elevated | | | | | | | | | | | |
| | Carbon Monoxide | | | Nitrogen | | | Carbon dioxide | | | Ethylene | | |
| | Dangerous | Elevated | Normal | Dangerous | Elevated | Normal | Dangerous | Elevated | Normal | Dangerous | Elevated | Normal |
| Dangerous | HR | HR | MR | HR | MR | MR | MR | MR | MR | HR | HR | MR |
| Elevated | HR | HR | MR | HR | MR | MR | MR | MR | LR | HR | HR | MR |
| Normal | MR | LR | LR | LR | LR | LR | LR | LR | LR | MR | LR | LR |

| Oxygen | Total Dissolved Combustible Gases (TDCG) | | | | | | | | | | | |
|---|---|---|---|---|---|---|---|---|---|---|---|---|
| | Normal | | | | | | | | | | | |
| | Hydrogen | | | Methane | | | Ethane | | | Acetylene | | |
| | Dangerous | Elevated | Normal | Dangerous | Elevated | Normal | Dangerous | Elevated | Normal | Dangerous | Elevated | Normal |
| Dangerous | HR | HR | MR | HR | HR | MR | HR | HR | MR | HR | HR | MR |
| Elevated | HR | HR | MR | HR | HR | MR | HR | HR | MR | HR | HR | MR |
| Normal | LR | LR | LR | LR | LR | LR | LR | LR | LR | LR | LR | LR |

| Oxygen | Total Dissolved Combustible Gases (TDCG) | | | | | | | | | | | |
|---|---|---|---|---|---|---|---|---|---|---|---|---|
| | Normal | | | | | | | | | | | |
| | Carbon Monoxide | | | Nitrogen | | | Carbon Dioxide | | | Ethylene | | |
| | Dangerous | Elevated | Normal | Dangerous | Elevated | Normal | Dangerous | Elevated | Normal | Dangerous | Elevated | Normal |
| Dangerous | HR | HR | MR | LR | MR | MR | MR | MR | MR | HR | HR | MR |
| Elevated | HR | HR | MR | LR | MR | MR | MR | LR | LR | HR | HR | MR |
| Normal | LR | LR | LR | LR | LR | LR | LR | LR | LR | LR | LR | LR |

## III. Results

FST was applied to ten bushings. The fuzzy rules were applied to each bushing. For each rule, the truth value of the consequence is the minimum membership value of the antecedent. The degrees of membership of the other gases are shown in Table 4.

### *A. Aggregated Rules*

The table of fuzzy rules can be simplified further by finding within compartments cells with common features. This process is called aggregating. One can develop the following aggregated rules (AR) based on the highlighted compartment in Table 2:

(AR4) IF bushing has 'Dangerous level of TDCG' AND 'NOT Normal Oxygen' AND 'Not Normal Methane', THEN the bushing belongs to 'Group A (high risk or dangerous)'.

(AR5) IF bushing has 'Dangerous TDCG' AND 'NOT Normal Oxygen' AND 'Normal Methane', THEN the bushing belongs to 'Group B (medium risk or elevated)'.

(AR6) IF bushing has 'Dangerous TDCG' AND 'Normal Oxygen' AND 'Not Normal Methane', THEN the bushing belongs to 'Group B (medium risk or elevated)'.

(AR7) IF bushing has 'Dangerous TDCG' AND 'Normal Oxygen' AND 'Normal Methane', THEN the bushing belongs to 'Group C (low risk or normal)'.

In rule AR1, the consequence is 'the bushing belongs to Group A'. The truth value of this consequence (CAR4) is shown in Equation (36).

$$CAR_4 = \min(1,1,1) = 1 \tag{36}$$

Where the values in the ***min*** function are obtained as follows: The first 1 is the degree of membership of TDCG for bushing#200323106 in the set 'Dangerous'. The second 1 is the degree of membership of NOT Normal Oxygen for bushing-200323106 is in the set 'NOT normal' is 1, which is obtained by subtracting the degree of membership of NOT Normal Oxygen in the set 'Normal', i.e. 0, from 1. The third 1 is the degree of membership of 'NOT Normal Methane' for bushing #200323106 in the set 'Normal' i.e. 0, from 1. Note that the 'NOT' operator requires that the corresponding degree of membership be subtracted from 1. An 'ANY' term entails summing of all the degrees of membership of a particular quality, e.g. acetylene, or TDCG in different corresponding sets. For an example, the condition 'ANY Level of TDCG' has a degree of membership of 1. This is obtained by summing the degrees of membership of TDCG for bushing #200323106 in the sets 'Dangerous' (1), 'Elevated' (0) and 'Normal' (0) as shown in Table 4.

Only the application of the rules in the highlighted compartment of Table 2 has been demonstrated. The application of the rules in all the other compartments follows the same pattern. Since different rules can result in the same conclusion or consequence, the truth values of a particular consequence will vary according to the rule applied to the bushing.

TABLE IV
MEMBERSHIPS OF GASES IN #200323106

| Gas | Quantity | Membership Fxn | Degree of Membership | Gas | Quantity | Membership Fxn | Degree of Membership |
|---|---|---|---|---|---|---|---|
| Acetylene | 0 | Normal<br>Elevated<br>Dangerous | 1<br>0<br>0 | Hydrogen | 5782 | Normal<br>Elevated<br>Dangerous | 0<br>0<br>1 |
| Carbon Dioxide | 72 | Normal<br>Elevated<br>Dangerous | 1<br>0<br>0 | Methane | 240 | Normal<br>Elevated<br>Dangerous | 0<br>0<br>1 |
| Carbon Monoxide | 44 | Normal<br>Elevated<br>Dangerous | 1<br>0<br>0 | Nitrogen | 4.58 | Normal<br>Elevated<br>Dangerous | 0<br>1<br>0 |
| Ethane | 22 | Normal<br>Elevated<br>Dangerous | 0<br>1<br>0 | Oxygen | 0.2535 | Normal<br>Elevated<br>Dangerous | 0<br>0<br>1 |
| Ethylene | 2 | Normal<br>Elevated<br>Dangerous | 1<br>0<br>0 | TDCG | 6090 | Normal<br>Elevated<br>Dangerous | 0<br>0<br>1 |

Once all the rules have been applied to a particular bushing, and different truth values of each consequence obtained, the ***maximum*** value of each consequence among all the rules that result in that consequence, is taken as the degree to which that consequence applies to a given bushing. This eventually gives rise to an aggregated fuzzy output as shown in Table 5 and Equation 37.

$$AGD_i = \max\left(CAR1_i \cap CAR2_i \cap \cdots \cap CARn_i\right) \tag{37}$$

Where

AGDi is the aggregated decision for category i, e.g. group HR, CARi is the consequence of aggregated rules in a particular category i, in a certain compartment. i is the number of categories, in this case the categories are HR, MR and LR.

TABLE V
AGGREGATED OUTPUT FOR BUSHING #200323106

| Category | Degree of membership |
|---|---|
| Group C (HR) | 1 |
| Group B (MR) | 1 |
| Group A (LR) | 1 |

According to Table 5, bushing #200323106 belongs to Group C (low risk or normal), it also belongs to Group B (elevated or medium risk) as well as Group A (high risk or dangerous) to degrees 1, 1 and 1, respectively. This step is the end of the fuzzification steps. The value of indicates that the degradation is severe but does not indicate how bad it is. It can be anywhere on the highlighted line in the MF curve. In the case of bushing #200323106, the position on the graph is insignificant because the value is already in the saturation region. But if the bushing had lower degrees of membership, (i.e. less than 1) in Group B and C, then one would be able to determine the extent of degradation and thus make an informed maintenance decision of a likely time to replace the bushing. The crisp result is useful for determining the degree of degradation, and not only informing of rejection or acceptance of a component. The crisp result takes into account the degrees of membership in all the groups in the decision truth table. In other words the crisp fuzzy result is useful for determining spares level at any given time and deciding the maintenance action. To quantify the extent of damage, the fuzzy information needs to be ranked to give crisp data.

### *B. Defuzzification*

Defuzzification is aimed at converting fuzzy information into crisp data. The method used for defuzzification in this case is called the weighted average of maximum values of membership functions method used by Siler [12] and Majozi [5]. The method was selected because it is effective and computationally inexpensive. The result from the application of this method gives the rank or level of risk of each bushing. For bushing #200323106 with an aggregated output is shown in Table 6, the rank is obtained using Equation (38). Figure 2 shows the aggregated membership function from which the values for Equation (38) are taken.

$$Rank(200323106) = \frac{\left(\frac{0+10}{2}\right)\cdot \mu(Group_A(x)) + 60\cdot \mu(Group_B(x)) + \left(\frac{80+100}{2}\right)\cdot \mu(Group_C(x))}{\mu(Group_A(x)) + \mu(Group_B(x)) + \mu(Group_C(x))} \quad (38)$$

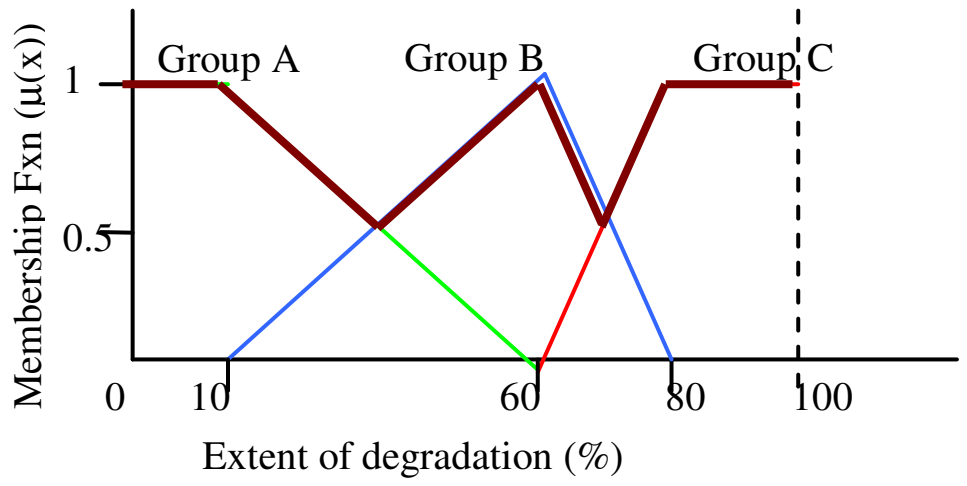

Fig. 2. Aggregated output for bushing #200323106

The coefficients appearing in Equation 38 are the levels of risk of failure corresponding to the maximum values, i.e. 1, of the respective sets as shown in the conclusion table, for example a risk of rating of 60 corresponds with the maximum value of the membership function of set B. In case there is a flat, as in the set A membership function as well as set C membership function, an average value of the extreme values at the maximum is used as a coefficient, e.g. (80+100). Thus the solution to (38) is shown in (39).

$$Rank(200323106) = \frac{\left(\frac{0+10}{2}\right)\cdot(1) + 60\cdot(1) + \left(\frac{80+100}{2}\right)\cdot(1)}{(1)+(1)+(1)} = 51.66 \quad (39)$$

Action is taken according to the crisp result. Bushings with a value of more than 30 are removed from service. Between 10 and 30 the interval of monitoring is halved or the frequency is doubled. Below 10 the bushing is left to operate as normal. Clearly bushing#200223106 with a crisp output value of 51.66 should be removed from service. Table 6 shows the results of all ten bushings.

Table 6 shows how a multiple layer perceptron (MLP) classified the same bushings that were used to demonstrate the application of fuzzy inference. The two methods show similar levels of accuracy. Manual evaluation of the crisp result from the fuzzy analysis showed no false acceptance rate, i.e. 100% accuracy. A neural network was able to classify the crisp data, using the criteria of x>30 for reject, x<30 for accept. Because the neural network in the second case used results from a fuzzy analysis it is called a neuro-fuzzy system. The neural network (NN) classified data directly from the DGA gas chromatography sheet [10] using a multiple layered perceptron with 7 hidden neurons, as done previously by Dhlamini and Marwala [11]. The manual method used an experienced maintenance operator, who is supposed to be 100% accurate. The results prove that NN and neuro-fuzzy have similar levels of accuracy (90%). While the purely fuzzy method showed 100% accuracy, NN are fast and efficient, taking 1.35s to train and classify the data compared to 30 minutes for the fuzzy set system and the neuro-fuzzy system, compared to 5 minutes for the manual method of classification of 10 bushings.

TABLE VI
CLASSIFICATION OF BUSHINGS

| Bushing | Rank | Fuzzy(manual) | NeuroFuzzy | NN | Human Decision |
|---|---|---|---|---|---|
| 200323106 | 51.666667 | Reject | Reject | Reject | Reject |
| 200373387 | 60 | Reject | Reject | Reject | Reject |
| 200323104 | 32.5 | Reject | Reject | Accept | Reject |
| 200323105 | 32.5 | Reject | Accept | Reject | Reject |
| 200302381 | 60 | Reject | Reject | Reject | Reject |
| 200355292 | 5 | Accept | Accept | Accept | Accept |
| 200367794 | 5 | Accept | Accept | Accept | Accept |
| 200378937 | 5 | Accept | Accept | Accept | Accept |
| 200328202 | 5 | Accept | Accept | Accept | Accept |
| 200365229 | 5 | Accept | Accept | Accept | Accept |
| | Accuracy | 100% | 90% | 90% | 100% |

## IV. CONCLUSION

The method of using fuzzy inference system (FIS) compares well with the method of diagnosing using neural networks. FIS tells maintenance personnel whether or not there is damage and also how severe is the damage, thus helping to make operational decisions if the bushings should be replaced or remain in service. The benefit of using FIS over neural networks is that it allows the user to evaluate the extent of damage more objectively and comprehensively. The crisp result from fuzzy analysis is useful for determining spares level at any given time and deciding the maintenance action. The neural network (NN) classified data directly from the DGA gas chromatography sheet using a multiple layered perceptron with 7 hidden neurons, as done previously by Dhlamini and Marwala. The manual method used an experienced maintenance operator, who is supposed to be 100% accurate. The results prove that NN and neuro-fuzzy have similar levels of accuracy (90%). While the purely fuzzy method showed 100% accuracy, NN are fast and efficient, taking 1.35s to train and classify the data compared to 30 minutes for the fuzzy set system and the neuro-fuzzy system, compared to 5 minutes for the manual method of classification of 10 bushings.